\documentclass{article}

\PassOptionsToPackage{numbers}{natbib}

\usepackage[final]{neurips_2020}

\usepackage[utf8]{inputenc} 
\usepackage[T1]{fontenc}    
\usepackage{hyperref}       
\usepackage{amssymb}
\usepackage{url}            
\usepackage{booktabs}       
\usepackage{amsfonts}       
\usepackage{nicefrac}       
\usepackage{microtype}      
\usepackage{xcolor}         
\usepackage{times}
\usepackage{amsmath, amssymb}
\usepackage{mathtools}
\usepackage{latexsym}
\usepackage{todonotes}
\usepackage{comment}
\usepackage{tabto}
\usepackage{soul}
\usepackage{multicol,multirow}
\usepackage{graphicx}
\usepackage{verbatim}
\usepackage{authblk}

\def\BibTeX{{\rm B\kern-.05em{\sc i\kern-.025em b}\kern-.08em
    T\kern-.1667em\lower.7ex\hbox{E}\kern-.125emX}}
    
\usepackage{color, colortbl}    
\usepackage{amssymb}
\usepackage{algorithm}
\usepackage[noend]{algpseudocode}
\usepackage{xcolor}

\author[1]{\textbf{Anastasiia Usmanova}}
\author[2]{\textbf{François~Portet}}
\author[2]{\textbf{Philippe Lalanda}}
\author[2]{\textbf{German Vega}}

\affil[1]{Univ. Grenoble Alpes, France}

\affil[2]{Univ. Grenoble Alpes, CNRS, Inria, Grenoble INP, LIG, 38000 Grenoble, France  \qquad}

\title{Federated Learning and catastrophic forgetting in pervasive computing: demonstration in
\newline
HAR domain}

\begin{document}

\maketitle

\begin{abstract}

Federated Learning has been introduced as a new machine learning paradigm enhancing the use of local devices.  At a server level, FL regularly aggregates models learned locally on distributed clients to obtain a more general model. In this way, no private data is sent over the network, and the communication cost is reduced. However, current solutions rely on the availability of large amounts of stored data at the client side in order to fine-tune the models sent by the server. Such setting is not realistic in mobile pervasive computing where data storage must be kept low and data characteristic (distribution) can change dramatically. To account for this variability, a solution is to use the data regularly collected by the client to progressively adapt the received model. But such naive approach exposes clients to the well-known problem of catastrophic forgetting. The purpose of this paper is to demonstrate this problem in the mobile human activity recognition context on smartphones. 
\end{abstract}
\medskip

\section{Introduction}

Pervasive computing promotes the integration of smart devices in our living spaces to develop services providing assistance to people \cite{becker}. For a decade, we have seen the emergence of smarter services based on Machine Learning (ML) and Deep Learning (DL) techniques. ML enables the production of highly performing decision systems by identifying patterns that may be hidden within massive data sets whose exact nature is unknown and therefore cannot be programmed explicitly. However, bringing such services into production raises a number of difficult issues. Current solutions rely on complex distributed architectures, integrating devices, edge, and cloud infrastructure. Simply put, learning models are built in the cloud using historical data, then deployed and executed on devices, or on the nearest edge resources. If possible, additional data is regularly collected by the devices and sent up to the cloud in order to build more up-to-date models. Such an approach, however, undergoes major limitations in terms of security (data sent over unprotected networks), performance (high volume of data), and cost related to communication and also models retraining (especially when using deep learning).

Federated Learning (FL) was recently proposed \cite{mcmahan} \cite{bonawitx} as a new learning paradigm promoting the execution and specialization of models on devices, called clients. At one point, clients models are sent to a server where they are aggregated into a more generic one. This new model is redistributed to the clients for a new local learning iteration, and so on. Federated Learning reduces communication costs and improves security because only models, and no private data, are exchanged \cite{8}. It has immediately attracted attention in the pervasive field but it rapidly appeared that adaptations were needed to meet specific challenges of this domain \cite{percom}. Originally, Federated Learning was set out as an alternative form of distributed learning to obtain a single well-performing model (on the server side) without ever communicating user data. However, to deal with pervasive computing, this learning approach should shift from a single server-centric model to a multi client-centric model objective \cite{lee2021opportunistic}. Precisely, client models should perform well on local data (\emph{strong personalization}) as well as on data only seen by others (\emph{good generalization}).

One particular thorny issue is when a client receives a new model from the server. This new model must be readjusted to the client in order to recover a good personalization, while preserving its recently obtained generalization. The solution commonly used today is to fine-tune (train) the received model with client data kept for this purpose. In practice, this requires a carefully selected set of representative training  data to obtain satisfactory performances \cite{percom}. However, such setting is not realistic in mobile pervasive computing where data storage must be kept low and data characteristic (distribution) can change dramatically. To account for this variability, a solution is to use the data regularly collected by the client to progressively adapt the received model. But such naive approach exposes clients to the problem of \emph{catastrophic forgetting} \cite{ewc}, which appears when a neural network is optimized on new data that are too different from the data previously used for training.

This is related to a more general neural network problem known as the \textit{stability-plasticity} dilemma \cite{stab-plas}, where plasticity and stability refer to integrating new knowledge and retaining previous knowledge respectively. Such dilemma is at the heart of the research area of Continual Learning (CL) also known as lifelong learning \cite{chen2018lifelong}.

\medskip

The purpose of this paper is to show that the problem of catastrophic forgetting is critical in a pervasive computing application using Federated Learning. It is demonstrated in the mobile Human Activity Recognition (HAR) domain. The paper is organized as it follows. In Section~2, we provide background on FL and CL and describe current solutions and issues. In Section~3, we present our experimental settings while section~4 demonstrates the addressed problem, that is catastrophic forgetting in Federated Learning. The paper is concluded by a synthesis and an outlook on future work.

\section{Background}

\subsection{Federated Learning}

Federated Learning promotes the computation of local models on devices and the aggregation of these models on a server to produce a new model that is sent back to clients for use and further training.
When the client model has been significantly updated or when a certain time has elapsed, the client model is sent to the server for a new aggregation (figure 1). A core incentive of FL in pervasive computing is to utilize mutual learning goals between clients to indirectly share  knowledge with one another. However, effective collaboration between unmoderated clients, is still a challenge. This is due to statistical heterogeneity (differences in individuals' usage) and system heterogeneity (difference between client data caused by different system traits). 

A number of studies have proposed intricate methods to mitigate the detrimental effects of heterogeneity. Specifically, we have identified methods such as FedProx \cite{li2020federated} and FSVRG \cite{konecny2016federated} which apply local model/gradient controlled learning that limits the divergence of clients. Another known method, as seen in FeSEM \cite{xie2020multicenter} and MLMG \cite{9431011} is to group clients or create more than one server model based on clients similarity.

\begin{figure}[!bh]
  \centering
  \includegraphics[width=\linewidth]{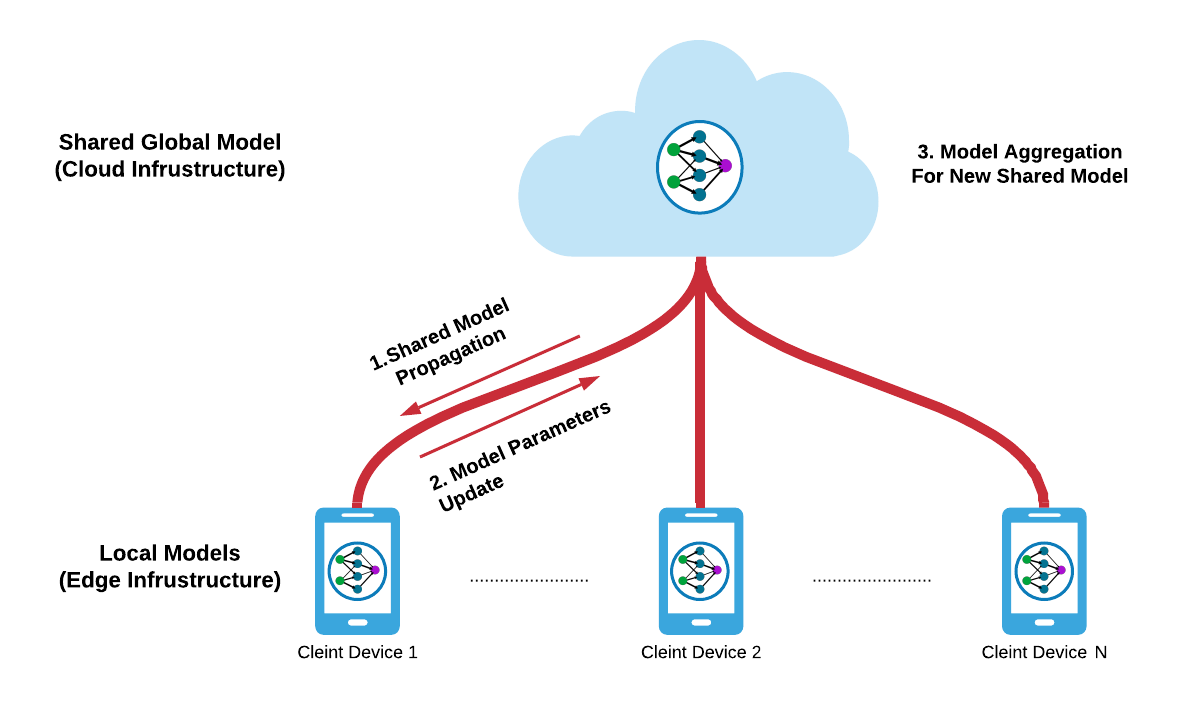}
  \caption{Federated Learning Architecture.}
  \label{fig:fl}
\end{figure}

A key point in FL is the way specialized models are aggregated. In the case of deep learning, two families of algorithms can be considered. The first strategy is to emphasize generalization. The aggregation algorithm considers local models in their entirety and builds a new model that potentially calls into question all layers and all weights associated with neurons. This approach is shown in FedAvg \cite{mcmahan2016communicationefficient}  and FedMA \cite{wang2020federated} algorithms. The second strategy focuses more on client specialization. Here, the algorithm does not question certain parts of the local models. Precisely, only the base layers are sent to the server for generalization, while the last layers are kept unchanged. This approach is used in FedPer algorithm \cite{fedperr}.

FL has been only recently used to implement pervasive services and, ultimately, does not address all the specifics of this domain well. The goal in pervasive computing, by contrast with the original FL, is to increase all individual clients performances, not only the server model performance. To do so, FL must account for the fact that data, even though in the same domain, might evolve very differently according to the clients. Also, mediation techniques \cite{mediation} \cite{mediation2} could be integrated into the FL approach in order to align the data in a more homogeneous way.

\subsection{Catastrophic Forgetting and Continual Learning}

Continual Learning (CL), also called lifelong learning or online machine learning, is the ability of a model to learn continually from an inﬁnite stream of data,  gradually integrating new acquired knowledge into old knowledge \cite{chenliu}.
Continual Learning highlights the sequential nature of the learning process. Formally, newly acquired data is passed to the model in the form of tasks. Tasks are made of labeled classes, uniquely identified and disjoint from each other, and a corresponding dataset. A model then learns sequences of tasks. At each training session, a model has only access to the current task (and possibly to some data kept from previous tasks), which is not accessible anymore after training.

In the past few years, numerous works have proposed to deal with Continual Learning and avoid catastrophic forgetting. They can be divided into three families \cite{class-inc}: \textit{regularisation-based} methods, \textit{exemplar-based} methods, and \textit{bias-correction} methods. Regularisation-based approaches minimize the impact of learning on the weights that are important for previous tasks. Among them, EWC \cite{ewc} calculates the weights of model parameters with a Fisher Information Matrix, while PathInt \cite{pathint} accumulates changes in each parameter during the learning process. RWalk \cite{rwalk} uses both  
approaches complemented with exemplars to improve results. Other approaches use knowledge distillation \cite{dist1} such as LwF \cite{lwf}  which trains a current model (student) with a past model as a teacher to prevent drifting while learning new tasks. The student model can also be guided using the attention of the teacher model as in LwM \cite{lwm}.

Exemplar-based methods, also called \emph{rehearsal methods}, rely on the storage of exemplars from previous tasks. For instance, iCarl \cite{icarl}, selects exemplars by their feature space representation and performs classification based on a nearest-class-mean rule in that feature space. EEIL \cite{eeil} uses an equal number of exemplars of all classes and a loss composed of a distillation measure to retain the knowledge acquired from the old classes, and a cross-entropy loss in order to learn the new classes.

Regarding the bias-correction approaches, we can mention  

BiC \cite{bic}, which uses an additional layer to correct task bias in the model and train the model in two stages. LUCIT \cite{lucir} uses a cosine normalization layer instead of the standard softmax layer and proposes to use margin ranking loss to prevent inter-task confusion. Finally, IL2M \cite{il2m} uses statistics of predictions of classes from previous tasks.

Overall, exemplar-based CL is the most efficient method but requires that a fair amount of data is kept on device, which is not possible when resources are limited. However, some works show that it is interesting to explore how a  small amount of exemplars can complement other techniques \cite{eeil}.

\section{Human Activity Recognition}

\subsection{Datasets and base model }\label{sec:har}

Experiments were performed in the domain of Human Activity Recognition (HAR) on smartphones. We believe that HAR is an excellent domain to test and better understand Federated Learning and Catastrophic Forgetting because activities tend to have generic patterns while being highly idiosyncratic. Collected data is very heterogeneous because it depends on people, devices, the way devices are carried, the environment, etc. Also, there are a number of realistic labeled datasets presenting heterogeneous data collected with different devices that can be used for our experiments. 

For the experiments, we have retained the UCI \cite{uci} dataset, which is heavily used by the HAR community as a benchmark. This dataset was collected with 30 volunteers using a Samsung Galaxy S II placed on the waist. One example of data contains 128 recordings of accelerometer and gyroscope (both 3 dimensions). There are 6 classes in the dataset (number of examples per class are in brackets):  0 -- Walking (1722), 1 -- Walking Upstairs (1544), 2 -- Walking Downstairs (1406), 3 -- Sitting (1777), 4 -- Standing (1906), 5 -- Lying (1944). 

HAR based on wearable sensors has prompted numerous research works, be they academic or industrial \cite{lara}. The most common approach is to process windows of data streams in order to extract a vector of features which, in turn, is used to feed a classifier. Today, the most popular and effective model is undoubtedly end-to-end deep neural networks. 

For that study, and all the experiments presented in this paper, we used the Convolution Neural Networks of \cite{percom} to take into account the small processing power on mobile devices. The studies \cite{percom} showed that a model 196-16C\_4M\_1024D gives the best result over a set of state-of-the-art-model. It includes 196 filters of a 16x1 convolution layer followed by a 1x4 max pooling layer, then by 1024 units of a dense layer and finally a softmax layer. 

We trained the model on the UCI dataset partitioned as follow: 70\% of data is in the train set, 15\% in the validation set, and 15\% in the test set. The model was trained using a mini-batch SGD of size 32 and a dropout rate of 0.5 and reached a 94.64\% accuracy on the testset. All experiments were written on Python 3 using the TensorFlow 2 library and run on a CPU Intel(R) Xeon(R) 2.30GHz (2 CPU cores, 12GB available RAM).

\subsection{PSA and T-SNE}\label{sec:pca}

To show how difficult it is for a model to distinguish the different classes in HAR, we made a Principal Component Analysis (PCA) \cite{pca} and a T-SNE analysis of the UCI dataset projected on a last layer (before activation) of the neural network.

\begin{figure}[!bh]
  \centering
  \includegraphics[width=\linewidth]{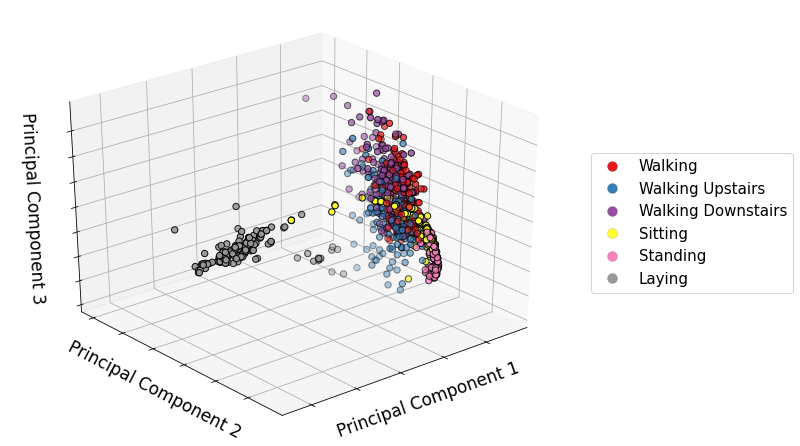}
  \caption{PCA of 200 examples of the UCI dataset on a last layer of the neural network (color display needed to read this figure).}
  \label{fig:pca}
\end{figure}

Figure \ref{fig:pca} shows the first three principal components of 200 randomly chosen examples of each class in a 3-dimensional space. We can see that the class \emph{Laying} is very distinguishable from the rest. \emph{Sitting} and \emph{Standing} are close to each  other. Walking movements are located in the same space, which shows they are hard to distinguish. 

\begin{figure}[!bh]
  \centering
  \includegraphics[width=\linewidth]{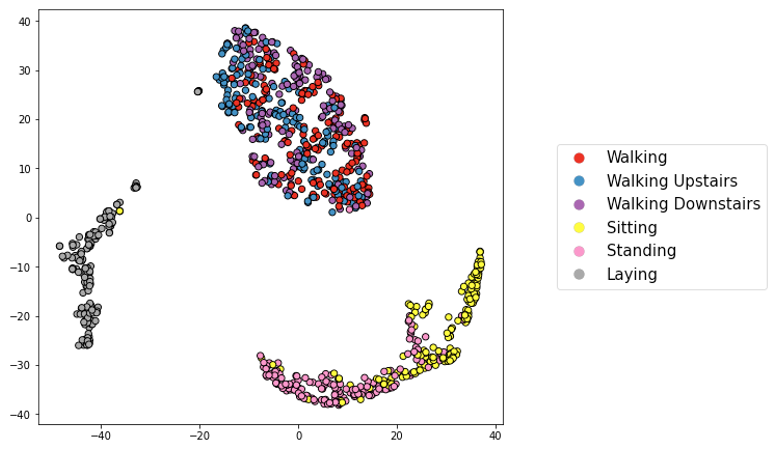}
  \caption{T-SNE of 200 examples of the UCI dataset on a last layer of the neural network (color display needed to read this figure).}
  \label{fig:tsne}
\end{figure}

Figure \ref{fig:tsne} presents the t-SNE analyzes of the dataset UCI HAR on a last dense layer before the ﬁnal activation of a pretrained CNN. T-Distributed Stochastic Neighbor Embedding (t-SNE) is a non-linear technique for dimension reduction by minimizing the divergence between two distributions: a distribution that measures pairwise similarities in high-dimensional space and a distribution that measures pairwise similarities of the corresponding low-dimensional space. It allows to ﬁnd patterns in the data by identifying observed clusters based on similarity of data points with multiple features. As t-SNE follows non-linearity, it can capture the structure of complex manifolds of higher-dimensional data and outperforms PCA, which is a linear feature extraction technique.

Again, we can see in this figure that it is very easy to distinguish the \emph{Laying} class from the other classes. The \emph{Sitting} and \emph{Standing} classes are located very close to each other. Movement-related classes are located in the same area, but they are individually hard to distinguish.

We will show in the following sections that such data distribution has a strong and direct impact on catastrophic forgetting.

\section{Demonstration of catastrophic forgetting}

\subsection{Experimental settings}

In order to demonstrate the problem of catastrophic forgetting, we have defined a Federated Learning architecture including a remote server and a number of clients simulated on laptops for reproducibility. Each client hosts a model to classify the UCI classes numbered from 0 to 5, i.e. \emph{Walking} is 0, \emph{Walking Upstairs} is 1, \emph{Walking Downstairs} is 2, \emph{Sitting} is 3, \emph{Standing} is 4 and \emph{lying} is 5. 

For the first learning round, all clients use the same neural network (the one defined to compute the PCA). For model aggregation at the server level, the FedAvg algorithm has been selected for its simplicity and good efficiency in the HAR domain \cite{percom}. Also, for the sake of clarity, all  the  clients behave synchronously, so all of them take part in each round.

At each FL communication round, each  client trains its local model on data from a sequence of disjoint tasks. A task is characterized by a set of classes and a dataset. Formally, each client $k \in \{ 1, 2, ..., K \}$  has its privately accessible sequence of $n_k$ tasks $\mathcal{T}_k$:
\vspace{-0.1cm} 
\begin{equation}
\mathcal{T}_k =  \left[\mathcal{T}_k^{1}, \mathcal{T}_k^{2}, ... , \mathcal{T}_k^{t}, ... , \mathcal{T}_k^{n_k}\right],
\mathcal{T}_k^{t} = (C^t_k, D^t_{k}),  \nonumber
\label{eq:task_seq_fcl}
\end{equation}

\noindent where $t \in \{1,...,n_k\}$, $C^t_k$ is a set of classes representing the task $t$ of a client $k$ (such that $C^{i}_k \cap C^{j}_k= \varnothing$ if $i \neq j$) and $D^t_k =  \{X^t_k,Y^t_k\}$ is training data corresponding to $C^t_k$.

\vspace{0.15cm}

A task can take place during several FL rounds. Formally, each task $\mathcal{T}_k^{t}$ for client $k$ is trained during $r^t_k$ communication rounds and $\sum_{t=1}^{n_k}r^t_k = R$, where $R$ is the total number of rounds between server and clients.

At each communication round, each client uses new training data, in order to simulate pervasive conditions of limited storage and incremental data collection. Data can also be new in the sense that it corresponds to a new task that has not been seen before by a client. Formally, at communication round $r$ client $k$ uses training data $D_{kr} = \{X_{kr}, Y_{kr}\}$:

\begin{equation}
D_{kr} = D^t_{kr} \subset D^t_k, \hspace{2mm}  \sum_{d=1}^{t-1}r^d_k < r \leqslant \sum_{d=1}^{t-1}r^d_k + r^t_k, \nonumber
\end{equation}

\noindent  where $D_{k'r'} \cap D_{k''r''} = \emptyset $, if $k' \neq k''$ and $r' \neq r''$

\noindent ($1 \leqslant k',k'' \leqslant K, 1 \leqslant r',r'' \leqslant R$);

\begin{figure*}[!ht]
    \centering
    \includegraphics[width=1
    \linewidth]{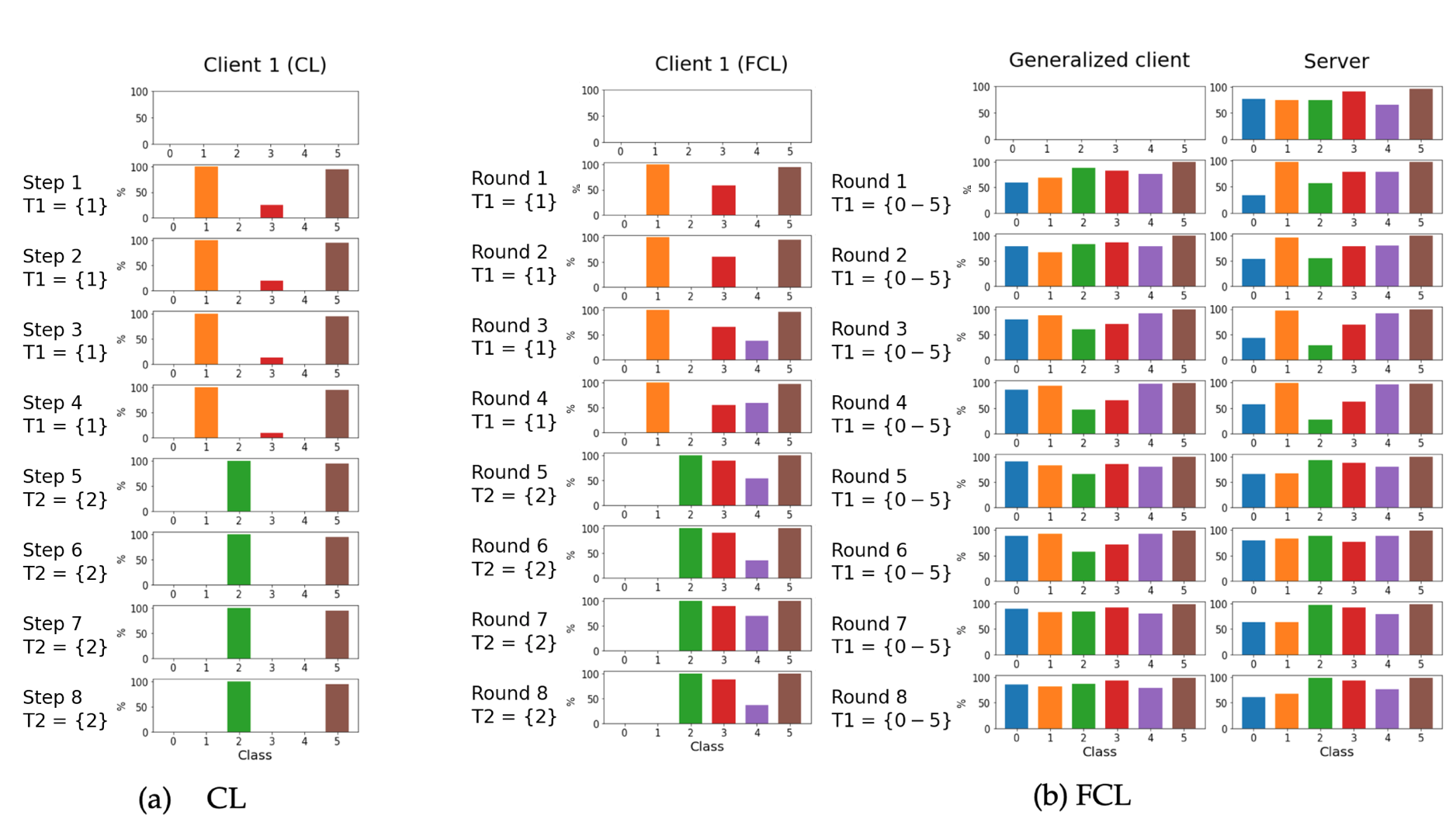}
    
    \vspace{-0.3cm}
    
    \caption{Demonstration of catastrophic forgetting in standard CL (a) and FCL (b).}%
    \label{fig:cat-for0}%
    
    \vspace{-0.4cm}
\end{figure*}

\vspace{0.1cm}

The notion of task is fundamental because it is the basis of the calculation of forgetting for clients.

\subsection{Baseline scenario}\label{sec:exp}

In order to have a baseline, we set up a first experiment with a single client, called \textbf{Client 1}. The purpose of this experience is to show how Continual Learning based on fine tuning in pervasive computing leads to catastrophic forgetting. 

The scenario implemented by this client is the following. Client 1 learns two successive tasks containing a single activity: \emph{Walking upstairs} and then \emph{Walking downstairs}. The client is trained alone iteratively for eight steps, four on the first activity and four on the second, with randomly chosen UCI data. Here, the steps are created to mimic the notion of rounds and ease the comparison with the FL experiment \emph{per se}.

\subsection{Federated Learning scenario}\label{sec:exp_FL}

For the second experiment, we integrated this client in a Federated Learning process, where  it  can  indirectly  benefit from knowledge acquired by other clients, and we defined a scenario very favourable to catastrophic forgetting. 

Precisely, we have defined a scenario based on eight FL communication rounds.  In the first four rounds, \textbf{client 1} learns a first task (\emph{Walking upstairs}). Then, in the next four rounds, it learns another task (\emph{Walking downstairs}). Formally, we can say that Client 1 learns $n_1 = 2$ tasks in total: $\mathcal{T}_1^1 = (C_1^1, D_1^1)$ and $\mathcal{T}_1^2 = (C_1^2, D_1^2)$, where $C^1_1 = \{1\}$ and $C^2_1 = \{2\}$; $\mathcal{T}_1 = \left[\mathcal{T}_1^{1}, \mathcal{T}_1^{2}\right]$,  $r_1^1=r_1^2=R/2$.  At each round, the model of this client is initialized with the model aggregated in the server and then trained with data collected between the communication rounds $D^t_{1r} \subset  D_1^t $

We assume $K-1$ other clients which only learn one task. These generic clients perform online-learning on the same well-balanced task containing all classes at each round. So, they learn $n_g = 1$ task in total: $\mathcal{T}_g^1 = (C_g^1, D_g^1)$, where $C^1_g = \{0,1,2,3,4,5\}$, and have their privately accessible sequence of tasks $\mathcal{T}_g = \left[\mathcal{T}_g^{1}\right]$, so $r_g^1=R$. In this task, all clients get the same number of examples (data) at each round, so the size of their dataset does not influence forgetting.

Hence, for the sake of simplicity, we assume that all other $K-1$ clients behave similarly, and that we can represent their influence in the FL process with a single \textbf{generalized client}. This is why, for aggregation at the server side, the weights of generalized and observed clients are $1/K * (K-1)$ and $1/K$ respectively. 

To build the train and test datasets, we randomly chose examples from the UCI dataset in accordance with the scenario. For each client $k$ and for each communication round $r$, we built a train set $D_{kr}$ of the same size. 

We computed the models performance on a common test set for all clients and the server. The test dataset includes 100 examples of each class (600 in total). We run 8 communication rounds and 10 epochs for the clients local training. We assumed that we have $K=5$ clients (the influence of $K-1$ of them represented by a single generalized client). The size of a dataset for each client $k$ and each round $r$ is $|D_{kr}| = 120$. We used a learning rate $\eta = 0.01$, dropout rate equal to 0.5, batch size $B = 32$ and SGD optimizer.  

\subsection{Results}

Figure~\ref{fig:cat-for0} shows the percentage of correctly classified examples from the test set after each of the eight steps/rounds. 

The left part of the figure is about the first experiment.  It appears that classes $C_0$, $C_2$, $C_3$, and $C_4$ are almost totally forgotten when \emph{Walking upstairs} ($C_1$) starts. Similarly, when the client moves on to  \emph{Walking downstairs} ($C_2$), $C_1$ is immediately forgotten. Only \emph{Sitting} ($C_5$), a static activity, is preserved even if there is no example of this class during training. As highlighted by the PCA, $C_5$ is very distinguishable from the others. The initially trained model successfully learnt it and the parameters used to classify it were not changed with further training on $C_1$ and $C_2$. 

The right part of the figure is about Federated Learning. It shows slightly better performances. It is apparent that FL successfully transferred to client $1$ the knowledge about static actions (classes $C_3$, $C_4$ and $C_5$) that has been learnt by other clients. Client $1$, which has never seen these actions, can recognize them with good accuracy. However, $C_2$ is forgotten by Client 1 when $C_1$ starts and, similarly, $C_1$ is totally forgotten by Client 1 when it moves on to $C_2$. Catastrophic forgetting is undeniable for those classes. 

$C_0$ (Walking) $C_1$ (Walking Upstairs) and $C_2$  (Walking Downstairs) are three very competitive classes as exhibited in Figure~\ref{fig:pca} and \ref{fig:tsne}. It seems thus that the absence of any of these classes will induce forgetting. This scenario is realistic since, in real-life, a person might spend much time walking without using stairs. During that time all information acquired in the model about stairs will be forgotten.

These experiments show that Federated Learning is clearly subject to the catastrophic forgetting issue when fine-tuning client models, although this problem can be attenuated by knowledge transfer between clients. However, with standard FL, this knowledge transfer also goes from Client 1 to the server model. It can be seen on Figure~\ref{fig:cat-for0} that the classes $C_0$, $C_1$ and $C_2$ of the server model have almost half the performances of the generalized client. Catastrophic forgetting has thus not only an effect on Client 1 but also on all the other clients.      
We are convinced that this forgetting problem is a real problem that current FL techniques cannot avoid in pervasive computing when data is heterogeneous, devices have limited resources. 


\section{Conclusion}

In this paper, we have highlighted in the Human Activity Recognition domain that the use of a Federated Learning does not avoid the catastrophic forgetting problem. Even if some knowledge, coming for various clients, is retained at the server level, some tasks can be very quickly forgotten.   This is due to the fact that some tasks are basically difficult to distinguish by a Neural Network (and they affect the same parts of a learning model). This is for example the case for movement activities in the HAR field. 

Dealing with catastrophic forgetting is then a necessary step for providing lifelong-learning in smart collection of objects. In our current work, we are investigating the use of distillation technique to transfer generic knowledge to specialized models. First results have shown this approach can greatly reduces the catastrophic forgetting effect \cite{anastasia}.

\section{Acknowledgements}
This work was partially supported by MIAI@Grenoble-Alpes (ANR-19-P3IA-0003).

\newpage

\bibliographystyle{plain}

{\footnotesize

\bibliography{comorea}}

\end{document}